\documentclass{article}
\usepackage{spconf,amsmath,graphicx}

\usepackage{enumitem}
\usepackage{subfigure}
\setlist{nosep, leftmargin=14pt}

\usepackage{mwe} 

\title{BEDs: Bagging Ensemble Deep Segmentation for Nucleus Segmentation with Testing Stage Stain Augmentation}
\name{\begin{tabular}{c}Xing Li $^{\star}$ $\quad$ Haichun Yang $^{\dagger}$ $\quad$ Jiaxin He $^{\star}$ $\quad$ Aadarsh Jha $^{\star}$ \\
\textit{Agnes B. Fogo} $^{\mathsection}$ $\quad$ \textit{Lee E. Wheless} $^{\mathparagraph}$ $\quad$ \textit{Shilin Zhao} $^{\dagger}$ $\quad$ \textit{Yuankai Huo} $^{\star}$\end{tabular}}

\address{$^{\star}$ Vanderbilt University, Computer Science, Nashville, TN, USA 37215  \\
$^{\dagger}$ Vanderbilt University Medical Center, Pathology, Microbiology and Immunology, \\Nashville, TN, USA 37232 \\
$^{\mathparagraph}$ Vanderbilt University Medical Center, Dermatology, Nashville, TN, USA 37232\\
$^{\mathsection}$ Vanderbilt University Medical Center, Biostatistics, Nashville, TN, USA 37232}

\usepackage{hyperref}
\begin{document}

\maketitle

\begin{abstract}
Reducing outcome variance is an essential task in deep learning based medical image analysis. Bootstrap aggregating, also known as bagging, is a canonical ensemble algorithm for aggregating weak learners to become a strong learner. Random forest is one of the most powerful machine learning algorithms before deep learning era, whose superior performance is driven by fitting bagged decision trees (weak learners). Inspired by the random forest technique, we propose a simple bagging ensemble deep segmentation (BEDs) method to train multiple U-Nets with partial training data to segment dense nuclei on pathological images. The contributions of this study are three-fold: (1) developing a self-ensemble learning framework for nucleus segmentation; (2) aggregating testing stage augmentation with self-ensemble learning; and (3) elucidating the idea that self-ensemble and testing stage stain augmentation are complementary strategies for a superior segmentation performance. Implementation Detail: \url{https://github.com/xingli1102/BEDs}.

\end{abstract}
\begin{keywords}
Nuclei segmentation, Deep learning, Bagging, Self-ensemble, Stain augmentation, Label fusion
\end{keywords}

\begin{figure}[htb]
\begin{minipage}[b]{1.0\linewidth}
  \centering
  \centerline{\includegraphics[width=8.5cm]{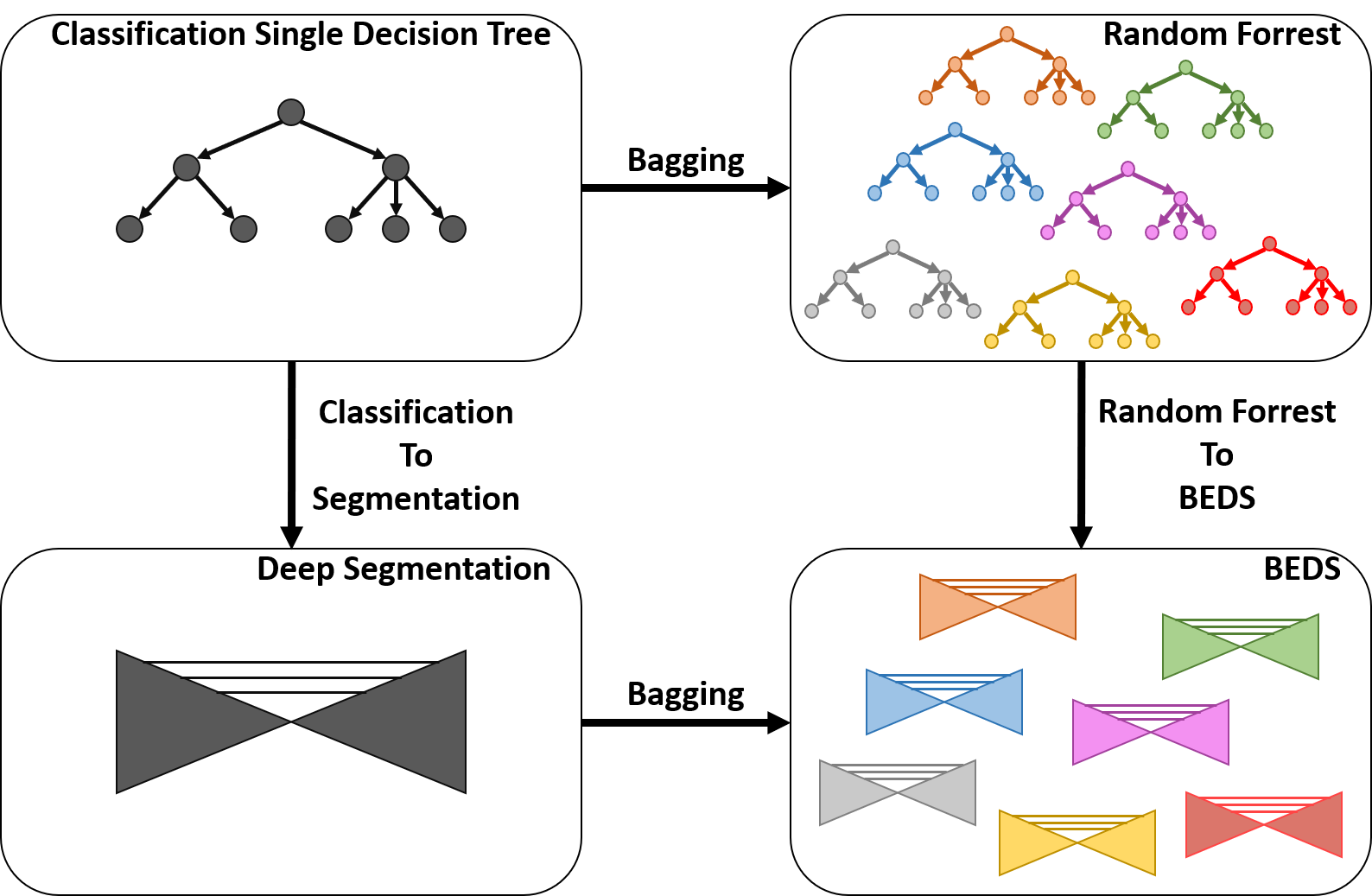}}
\end{minipage}
\caption{The upper row shows the evolution from a single decision tree to the random forest for classification. The bottom row shows the idea of using multiple segmentation models as the proposed BEDs algorithm, which is inspired by random forest algorithm.} \vspace{-3em}
\label{fig:Intro}
\end{figure}

\vspace{-0.5em}
\section{Introduction}
\label{sec:intro}
Cell nuclei segmentation on histopathological images is an essential task towards a computer-aided diagnosis system for supporting cancer diagnosis~\cite{cheng2017integrative}. The current deep learning-based approaches have shown their advantages in terms of higher accuracy and efficiency~\cite{MoNuSeg18, NoNewNet}. However, to automate cell nuclei segmentation is still a challenging task due to the high heterogeneity of shape, context and stain across different histopathological images. Many canonical machine learning strategies, such as data augmentation and ensemble learning, have been aggregated with deep learning to improve the generalizability~\cite{EmsembleMR, li2019u} .

Inspired by the random forest, we present a simple bagging ensemble deep segmentation (BEDs) method to aggregate deep segmentation networks with testing stage stain augmentation as shown in \textbf{Fig. \ref{fig:Intro}}. Briefly, 33 U-Nets were trained with partial training data to segment dense nuclei on pathological images, with 7 representative staining patterns during testing stage augmentation.

The contributions of this study are three-fold: (1) developing a self-ensemble learning framework for nucleus segmentation; (2) aggregating testing stage augmentation with self-ensemble learning; and (3) elucidating the idea that self-ensemble and testing stage stain augmentation are complementary strategies to achieve an overall superior segmentation performance. The proposed BEDs approach is adaptable framework, which is compatible with other segmentation networks and applications.

\section{Method}
\label{sec:method}

The proposed method (\textbf{Fig. \ref{fig:BEDs_Structure}}) aims to improve the robustness of nucleus segmentation by aggregating self-ensemble learning and testing stage augmentation. 

\vspace{-0.5em}

\subsection{Training Segmentation Models}
\label{ssec:method_data}

By following the random forest design \cite{randomforest}, $n$ subsets are randomly sampled from the original training dataset. Next, $n$ U-Net segmentation models, are trained with $n$ subsets, which play similar role as decision trees in a random forest. The U-Net is implemented based on~\cite{pix2pix}. In this study, $n$=33 subsets are randomly sampled from the original 1356 training images, where each subset has 904 images (2/3 of the whole dataset). The network structure of 33 U-Nets are kept the same, using the default implementation. The best model of each U-Net is selected based on the DicecSimilarity Coefficients (DSC) scores on the validation dataset. Based on our experience, the best model was achieved around $20^{th}$ epoch, which varied across different training subsets.

\vspace{-0.5em}

\subsection{Testing Stage Stain Augmentation}
\label{ssec:method_stain_augmentation}

After training $n$ models, we aggregate testing stage augmentation with bagging self-ensemble learning to further improve the performance. One critical uncertainty of segmenting nuclei on pathology images is the heterogeneous image appearance, even for the same Hematoxylin \& Eosin (H\&E) staining. Herein, we perform testing stage stain normalization to reduce the gap of appearance between testing and training images. 

\begin{figure}[t]
\begin{minipage}[b]{1.0\linewidth}
  \centering
  \centerline{\includegraphics[width=1.0\textwidth]{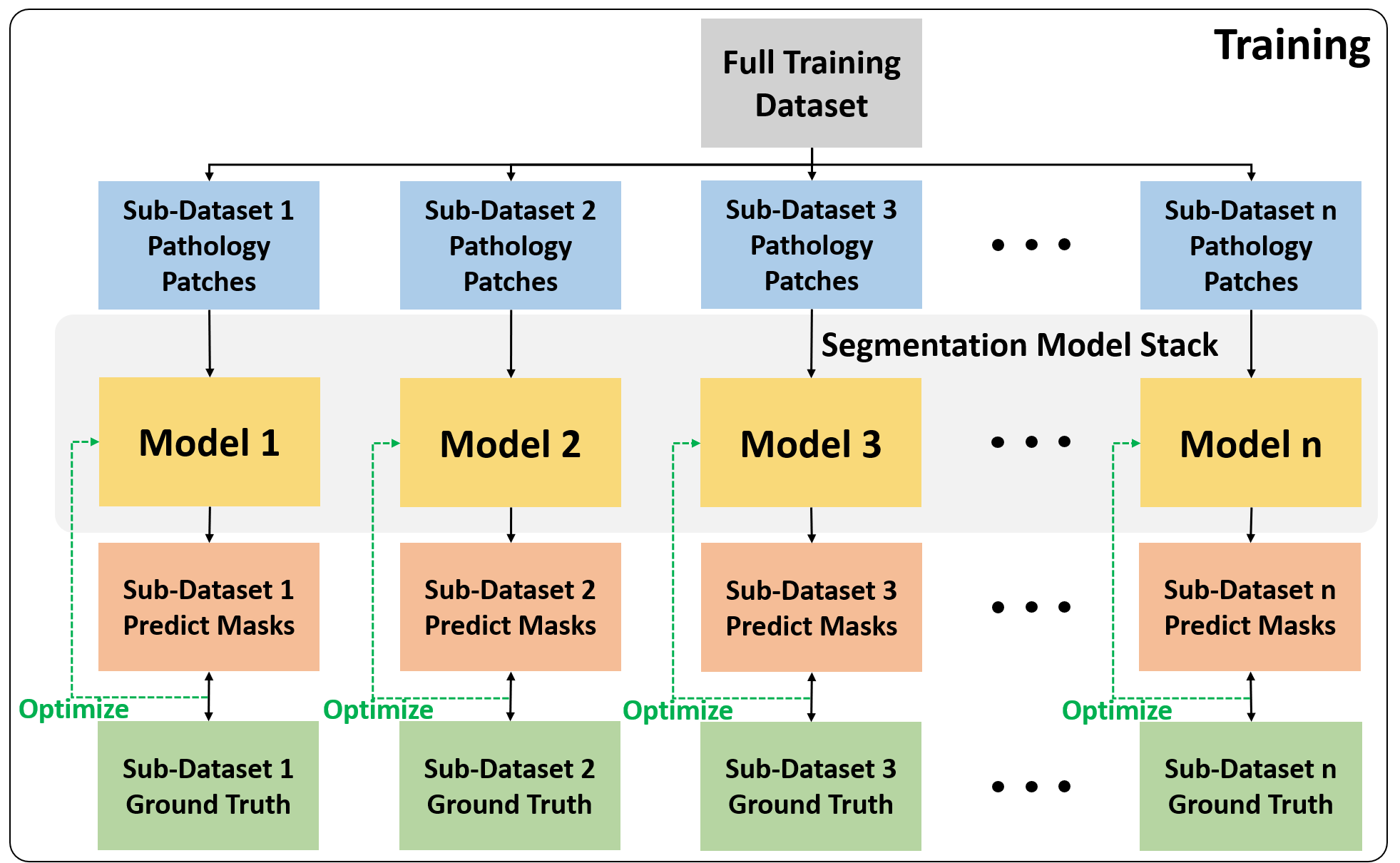}} \vspace{-0.2em}
  \label{fig:BEDs_Structure_a}
\end{minipage}
\begin{minipage}[b]{1.0\linewidth}
  \centering
  \centerline{\includegraphics[width=1.0\textwidth]{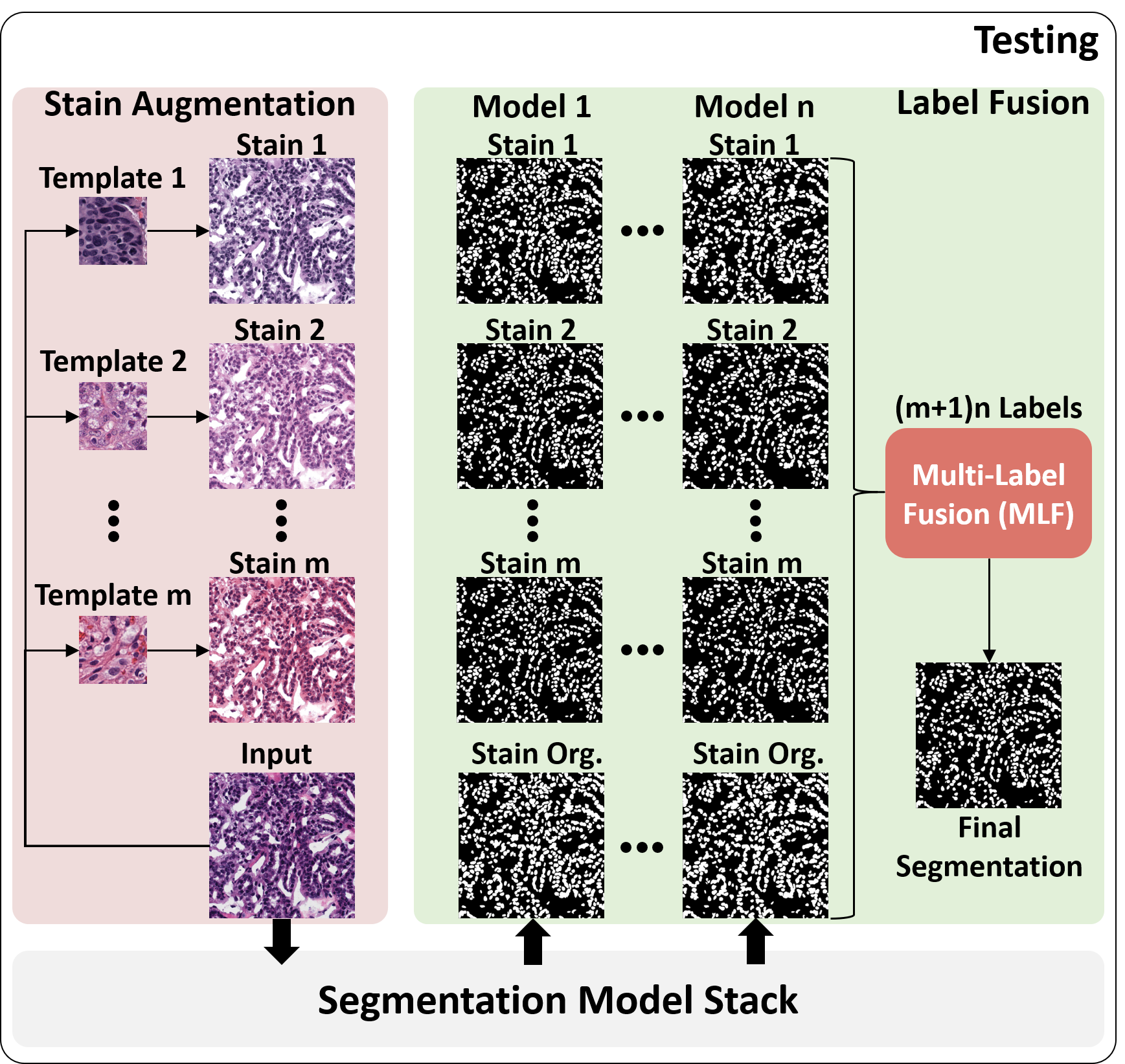}}\vspace{-0.2em}
  \label{fig:BEDs_Structure_b}
\end{minipage}
\caption{Shows the training and testing phases for the proposed method. The top figure shows the pipeline for training n models in the segmentation model stack. The bottom figure demonstrates testing stage label fusion.} \vspace{-1em}
\label{fig:BEDs_Structure}
\end{figure}

Stain normalization is typically performed between two image styles. However, since the training images are from heterogeneous resources and tissues, the variations cannot be defined as a single image style. Therefore, a data driven clustering procedure is employed to get the representative styles from the training data. First, low dimensional features of all training images are obtained from an ImageNet\cite{ImageNet} network, which is pretrained with ResNet-18 \cite{Resnet18}. Then a k-means algorithm is used to cluster the entire training data into $m$ clusters. Then, the $m$ images that are closest to the cluster centers are used as the templates to represent the heterogeneous appearances of training data (\textbf{Fig. \ref{fig:BEDs_Structure}}). $m$ is empirically set to 6 for all experiments in this paper.

With $m$ style templates, we perform $m$ stain normalization procedures between each testing image and template. As opposed to performing the augmentation in the training phase, stain normalization is performed during testing stage to harmonize the unseen H\&E staining patterns. Briefly, Vahadane Stain Extractor (VSE) \cite{VSE} tool is employed for stain normalization, which preserves biological structure by using color mixture modeling on sparse non-negative matrix factorization (SNMF). As suggested in \cite{VSE}, the hyper-parameter $\lambda$, which controls the trade-off between sparseness and the reconstruction accuracy, is set to 0.1. 

\begin{figure*}
    \centering
    \includegraphics[width=1.0\textwidth]{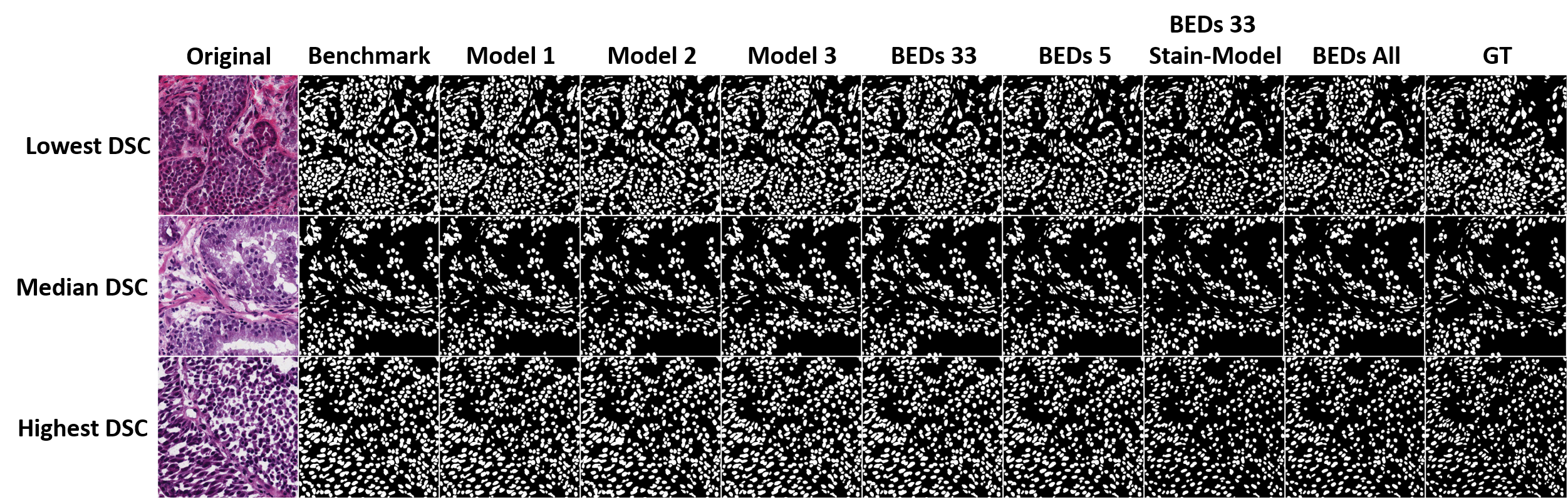}  \vspace{-1em}
  \caption{Shows the qualitative segmentation results across the different methods. The three rows show the testing images with lowest, median, and highest DSC using methods discussed in the experiments.} \vspace{-1em}
  \label{fig:mask_example}
\end{figure*}

\subsection{Label fusion}
\label{ssec:method_ensemble}
Since the training images have the same resolution of 256$\times$256 pixels, tiling is required to segment an arbitrary larger size testing image during inference. For any testing images, we tile them to 256$\times$256 pixels patches with overlapped regions (20 pixels width). After segmenting all patches, they are merged to original image scale by performing a bitwise-AND operation for the overlapped regions.\\
As shown in \textbf{Fig. \ref{fig:BEDs_Structure}}, stain augmentation is applied to each testing image. Next, $m$ stain normalized images along with the original image are fed to the segmentation model stack. In this study, $(m+1)n$ segmentation masks are fused using a pixel-wise majority vote (MV) algorithm:
\begin{equation}
    X^*(i)=
    \begin{cases}
    1,& \text{if } \sum_{p=1}^{m+1} \sum_{q=1}^{n} X_{p,q}(i)>\frac{(m+1)n}{2}\\
    0,              & \text{otherwise}
    \end{cases}
\end{equation}
where $X^*(i)$ is the final binary segmentation mask at pixel $i$, and $X_{p,q}(i)$ is the mask prediction of the $p^{th}$ stain style and the $q^{th}$ model at pixel $i$.

\vspace{-1em}
\section{Data}
\label{sec:data}
 A public dataset with 1356 annotated image patches (14 cancer types)~\cite{trainingData} was used as a training dataset. Each training patch had a resolution of 256$\times$256 pixels. 30 images and 14 images from the Multi-organ Nuclei Segmentation (MoNuSeg2018) challenge~\cite{MoNuSeg18} were used as validation and testing data, respectively. Both validation and testing sets had the image resolution of 1000$\times$1000 pixels. This research study was conducted retrospectively using human subject data made available in open access by~\cite{trainingData,MoNuSeg18}. Ethical approval was not required as confirmed by the license attached with the open access data.
 \vspace{-1em}

\begin{table}[h]
\caption{Pixel-wise DSC and object-wise F1 results.}
\centering
\begin{tabular}{c|cc|c}
\hline
Exp. & Mean & Median & F1\\
\hline
Benchmark  & 0.7959 & 0.7987 & 0.8702 \\
Model 1 & 0.7943 & 0.7960 & 0.8662 \\
Model 2 & 0.7934 & 0.7977 & 0.8695 \\
Model 3 & 0.7953 & 0.7979 & 0.8720 \\
BEDs 5 & 0.8106 & 0.8123 & 0.8830 \\
BEDs 33 & 0.8051 & 0.8089 & 0.8779 \\
BEDs 33 Model-Stain & 0.8093 & 0.8161 & 0.8762 \\
BEDs 33 Stain-Model & 0.8089 & 0.8152 & 0.8743 \\
BEDs 33 All & \textbf{0.8177} & \textbf{0.8192} & \textbf{0.8836} \\
\hline
\end{tabular}\vspace{-1em}
\label{tab:stats}
\end{table}

\section{experiments and results}
\label{sec:experiments}

The following experiments were performed on a workstation with 1 GTX-1080Ti GPU to evaluate the segmentation performance. 

\noindent\textbf{Benchmark:} The baseline segmentation model was a single U-Net trained by the entire training dataset with 1356 images.

\noindent\textbf{Model 1,2,3:} As each model in BEDs was trained by a random 904 images (2/3) from the entire training pool, we randomly picked three models from all 33 models as Model 1, Model 2, and Model 3 to show the variations of individual segmentation models.

\noindent\textbf{BEDs 5 Model:} This represented the segmentation results of assembling five randomly selected models from 33 models.

\noindent\textbf{BEDs 33 Model:} This represented the segmentation results of assembling all 33 models.

\noindent\textbf{BEDs 33 Model-Stain:} First, segmentation results from all 33 models for each stain augmentation were fused by a majority vote label fusion. Then, the fused segmentation results of 7 stains were further fused by performing another majority vote.

\noindent\textbf{BEDs 33 Stain-Model:} This was similar as the BEDs 33 Model-Stain, but we fused the results from 7 stain augmentations first.

\begin{figure}[h]
  \centering
 \includegraphics[width=8.5cm]{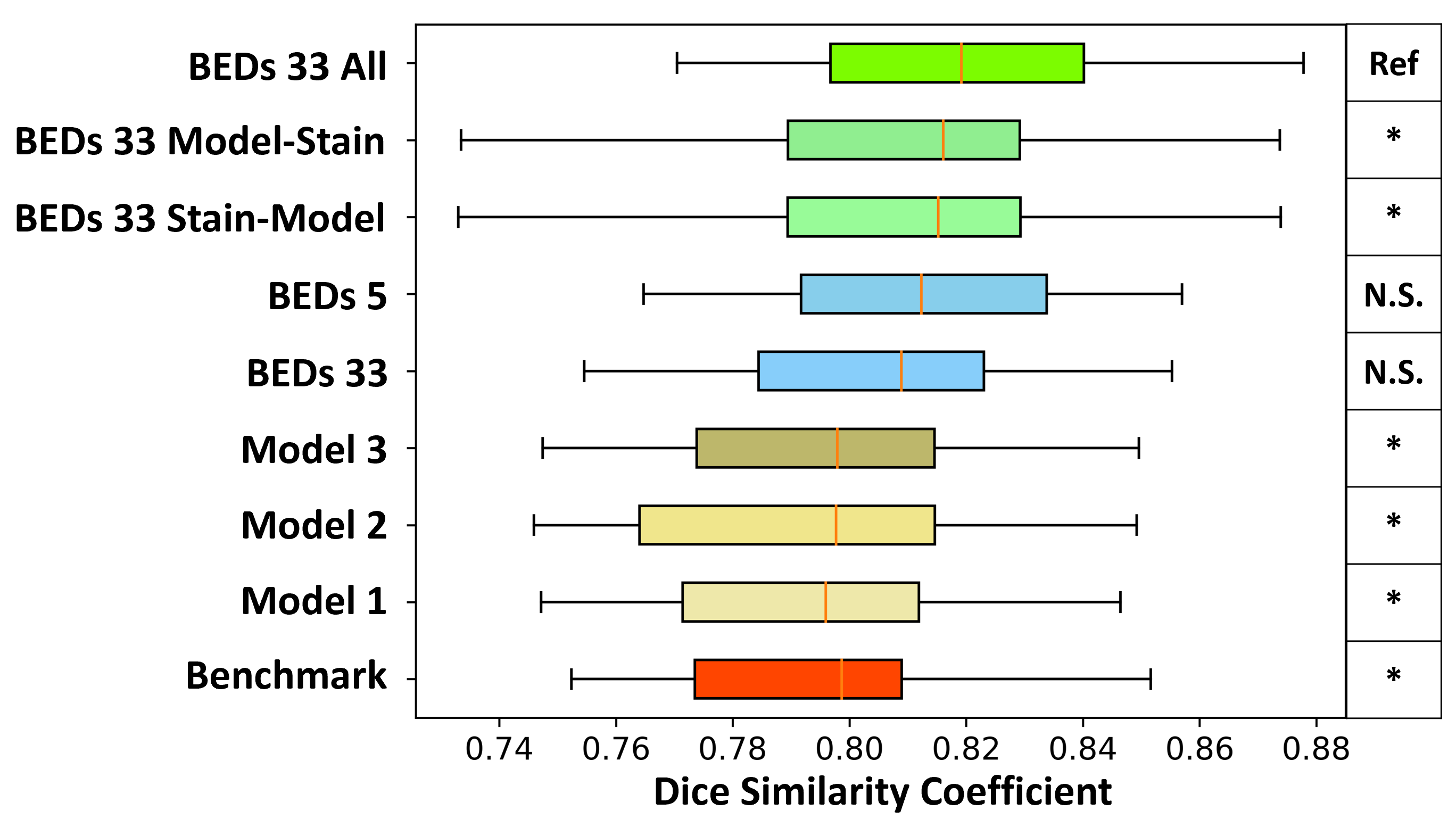}
\caption{Shows the quantitative results for all experiments. BEDs 33 All yielded the best performance in terms of average DSC, range, and also standard deviation. The right column shows the Wilcoxon signed-rank test results by using BEDs 33 All as reference. \textbf{*} represents significant (p$<$0.05), while \textbf{N.S.} stands for not significant.}
\vspace{-0.5em}
\label{fig:boxplot}
\end{figure}

\noindent\textbf{BEDs 33 All:} This was fused by all 33$\times$7 results simultaneously using the majority vote algorithm.

\begin{figure}[t]
  \centering
\includegraphics[width=8.5cm]{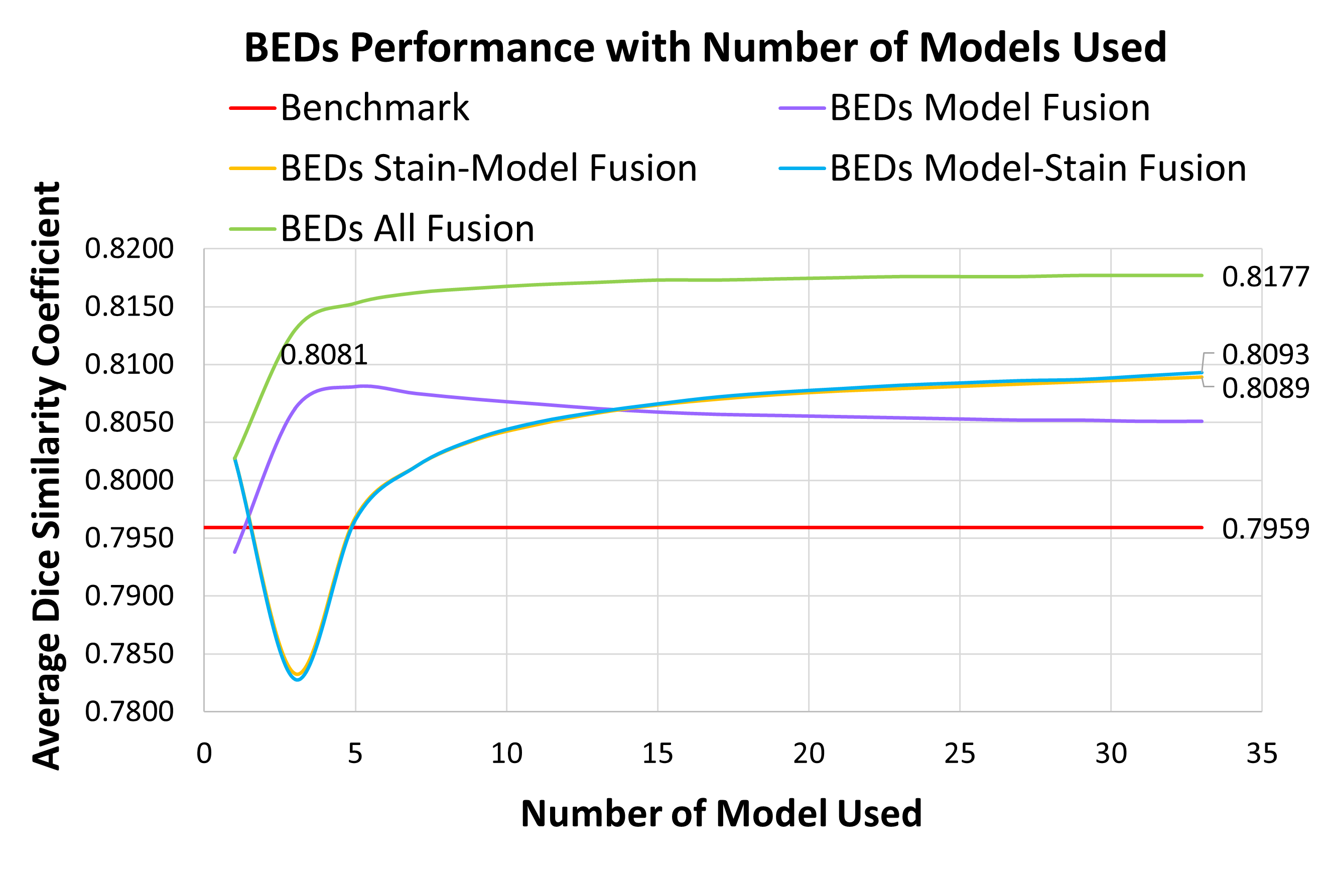}
\vspace{-1.5em}
\caption{This figure shows the ablation study of different ensemble strategies with different $n$ number of U-Nets. The red line is the benchmark, which is a single U-Net trained by all training data. From the results, the self-ensemble learning (without testing stage stain augmentation) achieved the best performance with a smaller $n$, while the testing-stage stain augmentation achieved the better performance with larger $n$. The holistic BEDs All model achieved the best performance since the self-ensemble learning and testing stage stain augmentation provided complementary improvements.}
\vspace{-1em}
\label{fig:Model_number_Compare}
\end{figure}

The qualitative results are shown in \textbf{Fig. \ref{fig:mask_example}}, while the quantitative results are presented in \textbf{Fig. \ref{fig:boxplot}} and \textbf{Table. \ref{tab:stats}}, where the pixel-wise DSC and object-wise F1 are computed. For the object-wise F1, a watershed instance segmentation algorithm was applied to both prediction and ground truth, and an Intersect-over-Union (IoU) threshold of 0.5 is set for true positive objects. \textbf{Fig. \ref{fig:boxplot}} also provides the Wilcoxon signed-rank test. An ablation study of different ensemble strategies with distinct $n$ number of U-Nets is shown in \textbf{Fig. \ref{fig:Model_number_Compare}}. The average run-time of a 1000$\times$1000 pixels image for Benchmark, Model 1, 2, and 3 is 1.27 sec, for BEDs 5 is 6.22 sec, for BEDs 33 is 37.98 sec, for all others is 38.78 sec. To sum, the BEDs 33 All models achieved the best accurarcy, while BEDs 5 is a compromise solution considering both computational efficiency and accuracy.

\vspace{-1em}
\section{Conclusions}
\vspace{-1em}
\label{sec:conclusion}
In this paper, we propose a self-ensemble learning algorithm with testing stage stain augmentation for nuclei segmentation. The ablation study in \textbf{Fig. \ref{fig:Model_number_Compare}} showed that the self-ensemble learning and testing stage stain augmentation were mutually complementary. Herein, the holistic model achieved the highest mean and median DSC, without using any extra training data.   

\vspace{-0.5em}

\section{Compliance with Ethical Standard}
\label{sec:ethic}
This research study was conducted retrospectively using human subject data made available in open access by~\cite{trainingData,MoNuSeg18}. Ethical approval was not required as confirmed by the license attached with the open access data. There is no conflicts of interests of all authors.

\vspace{-0.5em}
\bibliographystyle{IEEEbib}
\bibliography{strings,refs}

\end{document}